\documentclass[conference]{IEEEtran}
\IEEEoverridecommandlockouts
\usepackage{cite}
\usepackage{amsmath,amssymb,amsfonts}
\usepackage{algorithmic}
\usepackage{graphicx}
\usepackage{textcomp}
\usepackage{xcolor}
\usepackage{booktabs}
\usepackage{multirow}
\usepackage{url}
\usepackage{balance}

\begin{document}

\title{\LARGE \bf
AquaBEV-Nav: Learned BEV Occupancy for Underwater Navigation and Exploration}

\author{\IEEEauthorblockN{Anonymous Author(s)}
}

\author{Trung Tien Dong$^{1*}$, Zhenqi Wu$^{1*}$, Sahasra Kondapalli$^{1}$, Jiayi Wu$^{3}$, Yi Sheng$^{2}$ and Xiaomin Lin$^{1}$%
\thanks{$^{*}$Equal contribution}
\thanks{This research was supported through a cooperative agreement between NOAA’s Office of Coast Survey and the University of South Florida through the Center for Ocean Mapping and Innovative Technologies (COMIT 2.0), NA26NOSX402C0001 and NVIDIA academic grant}%
\thanks{$^{1}$ERA Lab, University of South Florida, Florida, USA. Emails: \texttt{\{dongt, zhenqi, xlin2\}@usf.edu}.}%
\thanks{$^{2}$YES Lab, University of South Florida, Florida, USA}%
\thanks{$^{3}$ University of Maryland- College Park, College Park, Maryland}
}

\maketitle

\begin{abstract}
Safe underwater exploration requires a robot to understand where surrounding
structures are located and which regions are available for motion. Existing
vision-based underwater exploration systems commonly obtain this information
indirectly by estimating monocular depth, unprojecting the geometry into 3D
space, and accumulating it into a 2D bird's-eye-view occupancy map. This reliance
on intermediate depth estimation is particularly problematic underwater, where
scattering and wavelength-dependent attenuation degrade
visual cues and limit the reliability of monocular depth estimates. We introduce
\textbf{AquaBEV-Nav}, an underwater exploration framework that bypasses explicit
monocular depth estimation through direct bird's-eye-view occupancy prediction.
Built upon the CORAL hierarchical exploration framework, AquaBEV-Nav replaces its
depth-based perception front end with AquaBEV. Given a single RGB frame, AquaBEV
maps visual features into a learned polar representation, performs causal
reasoning along the range dimension, and reconstructs local Cartesian occupancy
without relying on intermediate depth prediction. The resulting occupancy map is
accumulated into CORAL's persistent spatial memory, providing spatial context for
VLM-based high-level planning and collision constraints for dynamics-aware local
trajectory generation. Across ten simulated reef environments and six occupancy
backbones evaluated under a single protocol, AquaBEV-Nav reaches 37.48 structure IoU and 53.2 target IoU, 88.95\% closed-loop coverage with zero collisions. 
\end{abstract}

\section{Introduction}

Underwater exploration, ecological monitoring, and infrastructure inspection require autonomous robots to reason about the spatial structure of their surroundings in order to navigate safely~\cite{mccammon2026coralhotspots,yuan2023marine,nauert2023inspection}. Beyond recognizing objects or semantic regions, a robot must determine which parts of the environment are occupied, which remain traversable, and how this information changes as the vehicle moves. Recent vision-language exploration systems~\cite{wu2025dream,wu2026coral} maintain this spatial understanding in a persistent bird's eye view (BEV) occupancy map, enabling a high level planner to reason over accumulated observations and generate collision free waypoints for low level tracking.

BEV occupancy has been widely studied in autonomous driving, where it provides a common spatial representation for perception and planning~\cite{li2022bevformer,huang2024gaussianformer}. AquaBEV~\cite{dong2026aquabev} extends this idea to underwater perception by predicting local BEV occupancy directly from a single RGB image using 3D imaging sonar supervision. However, AquaBEV is optimized as a perception model rather than as the spatial representation consumed by a navigation system. Current underwater navigation systems instead commonly construct persistent maps through monocular depth estimation, 3D unprojection using camera geometry and vehicle pose, and rasterization into a 2D occupancy grid. Errors introduced during depth estimation and geometric reconstruction can therefore propagate into the map that ultimately determines navigation behavior.

\begin{figure}[t]
    \centering
    \includegraphics[width=1.0\columnwidth, trim=0 0 0 0, clip]{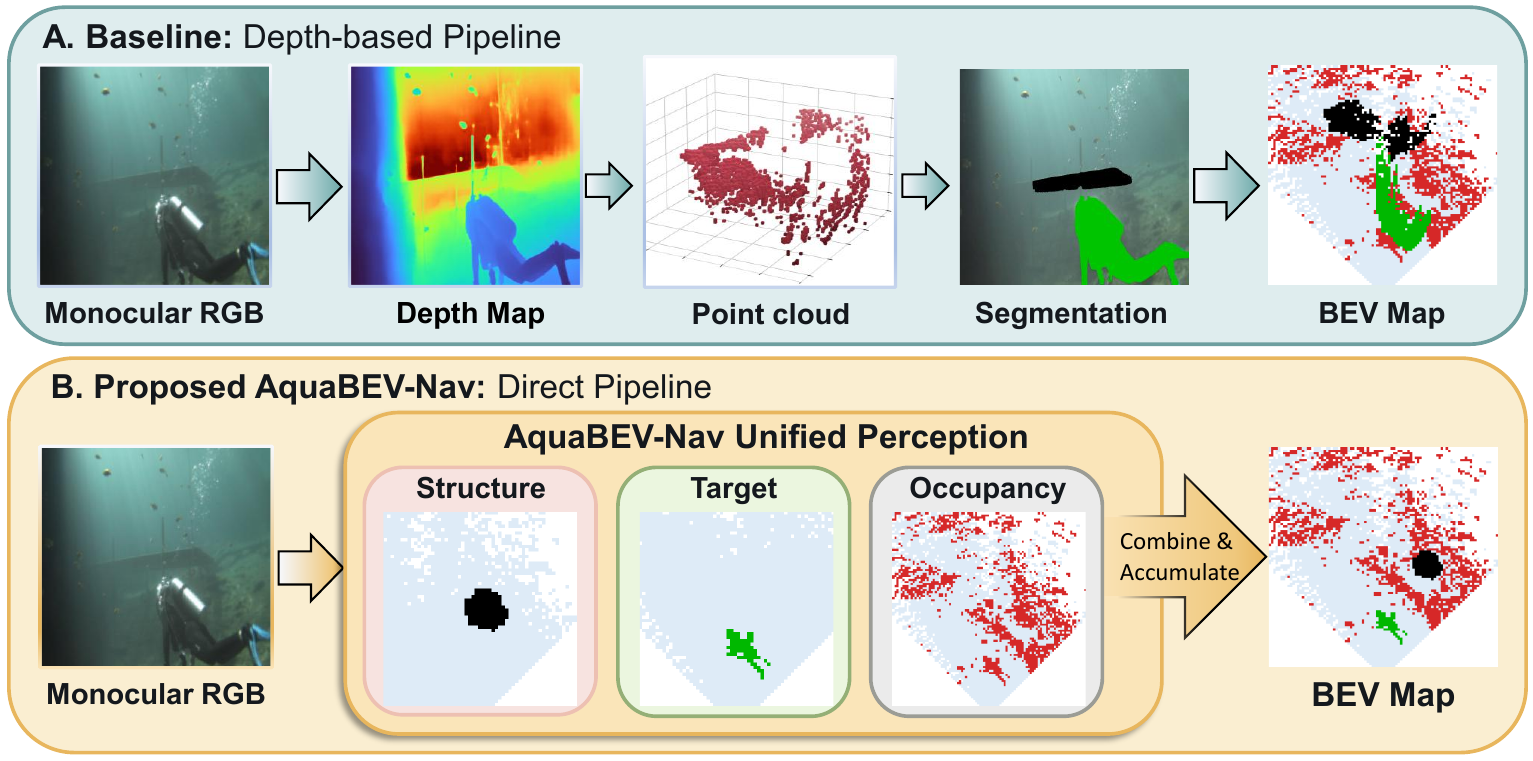}
    \vspace{-3mm}
    \caption{\textbf{AquaBEV-Nav.}
    CORAL constructs occupancy through monocular depth estimation, geometric reconstruction, and rasterization (top).
    AquaBEV-Nav replaces this perception pipeline with direct monocular BEV occupancy prediction (bottom).}
    \label{fig:pipeline}
    \vspace{-4mm}
\end{figure}

This observation motivates \textbf{AquaBEV-Nav}. Rather than estimating depth as an intermediate representation, AquaBEV-Nav connects learned BEV occupancy directly to the navigation stack, as showm in Fig. \ref{fig:pipeline}. Direct substitution, however, is not sufficient. Navigation needs more than a single occupancy field, it needs obstacles localized accurately enough to avoid, and targets localized accurately enough to approach. Occupancy predictors are optimized for perception metrics such as IoU or geometric agreement, which score a cell as right or wrong regardless of how far it lies from the truth, whereas navigation is sensitive to spatial error in proportion to its effect on traversability and collision margin. We analyze how learned occupancy errors propagate into navigation, decompose the predicted field into the obstacle and target evidence the planner consumes, and introduce spatially calibrated BEV objectives that align prediction with downstream navigation

Our contributions are:
\begin{itemize}
    \item We introduce \textbf{AquaBEV-Nav}, which connects monocular BEV occupancy directly to hierarchical underwater navigation.

    \item We develop spatially calibrated BEV objectives derived from the observed navigation relevant error structure, improving the planning representation without increasing model parameter count.

    \item We conduct a controlled evaluation across six occupancy backbones in simulation and physical pool experiments to study how occupancy prediction quality translates into downstream navigation performance.
\end{itemize}
\section{Related Work}

\subsection{Underwater Navigation and Spatial Mapping}

Underwater navigation requires a robot to reason about surrounding geometry despite the absence of GPS, degraded visual observations, and constrained vehicle dynamics~\cite{paull2014auvnav}. Classical autonomy has addressed these challenges through explicit mapping and motion planning, including obstacle aware planning~\cite{xanthidis2020obstacles}, perception aware trajectory generation~\cite{xanthidis2021aquavis}, and adaptive exploration of unknown environments~\cite{girdhar2014exploration}. More recent systems increasingly connect learned perception with navigation. Semantic scene understanding has been used to guide autonomous cave exploration~\cite{gupta2025cavepi}, while information driven navigation has been learned from demonstrations~\cite{lin2024uivnav}. Across these approaches, navigation depends on maintaining a spatial representation that extends beyond the current observation and allows the robot to reason about where it can move.

Foundation models have expanded this direction by providing semantic reasoning and higher level goal selection. Language grounded maps~\cite{huang2023vlmaps} and semantic frontier exploration~\cite{chaplot2020semexp} established similar ideas in embodied navigation, while recent work has begun transferring them to marine environments~\cite{yang2024oceanplan,saad2025aquachat}. However, vision language models still have limitations in precise spatial grounding~\cite{chen2024spatialvlm} and long horizon planning~\cite{kambhampati2024llmmodulo}. As a result, recent systems commonly combine semantic reasoning with an explicit geometric representation used for verification and local planning. CORAL~\cite{wu2026coral} follows this structure by maintaining persistent occupancy that supports waypoint generation and collision avoidance. These systems establish the importance of spatial maps for underwater autonomy, but the construction and quality of the occupancy representation are generally not their primary focus.

\subsection{BEV and Occupancy Prediction}

Bird's eye view occupancy provides a compact representation of surrounding free, occupied, and unknown space in a common coordinate frame \cite{dong2026post}. Lift Splat Shoot~\cite{philion2020lss} transports image features into BEV through predicted depth distributions, while BEVDepth~\cite{li2023bevdepth} improves this process using explicit depth supervision. Query based approaches instead learn the transformation between perspective observations and spatial representations. BEVFormer~\cite{li2022bevformer} aggregates image features into learned BEV queries, while later methods represent scene occupancy through orthogonal planes~\cite{huang2023tpvformer}, multi scale volumetric reasoning~\cite{wei2023surroundocc}, or Gaussian primitives~\cite{huang2024gaussianformer,huang2025gaussianformer2}. These methods demonstrate that useful spatial representations can be learned without requiring a conventional depth map as the final perception output.

Polar representations provide another direction for spatial prediction when measurements are organized by bearing and distance. PolarNet~\cite{zhang2020polarnet}, PolarFormer~\cite{jiang2023polarformer}, and PVP~\cite{xue2025pvp} show that polar coordinates can better reflect radial sensing geometry and range dependent spatial structure. Most BEV and occupancy methods, however, have been developed for terrestrial sensing with well characterized camera geometry, dense geometric supervision, and substantially larger training than are typically available underwater. AquaBEV~\cite{dong2026aquabev} adapts this family to underwater perception by using paired 3D imaging sonar as geometric supervision for monocular BEV occupancy. 

\subsection{From Occupancy Prediction to Navigation}

Although learned occupancy provides the representation required by a planner, occupancy prediction and navigation are commonly evaluated with different objectives. Perception methods emphasize overlap and geometric accuracy~\cite{huang2024gaussianformer,dong2026aquabev}, while navigation systems evaluate quantities such as collision avoidance, coverage, and path efficiency~\cite{katyal2021high,wu2026coral}. This separation leaves the relationship between occupancy quality and navigation behavior less understood.

A map with strong aggregate occupancy accuracy is not necessarily equally useful for planning. False occupied regions can remove traversable space, while false free regions can increase collision risk. Prior work on risk aware occupancy mapping and predicted occupancy navigation has shown that map uncertainty and occupancy structure directly affect path selection and navigation safety~\cite{laconte2021risk,katyal2019uncertainty}. These errors can become more consequential when local predictions are accumulated into persistent spatial memory and influence later planning decisions~\cite{wu2026coral}. AquaBEV-Nav studies this interface by inserting learned occupancy into CORAL while keeping the downstream mapping, planning, and control components fixed. Rather than asking only whether monocular occupancy is accurate, we study whether its spatial structure is suitable for the navigation system that consumes it.

\section{AquaBEV-Nav}
\label{sec:method}

\begin{figure*}[t]
    \centering
    \includegraphics[width=1.0\textwidth]{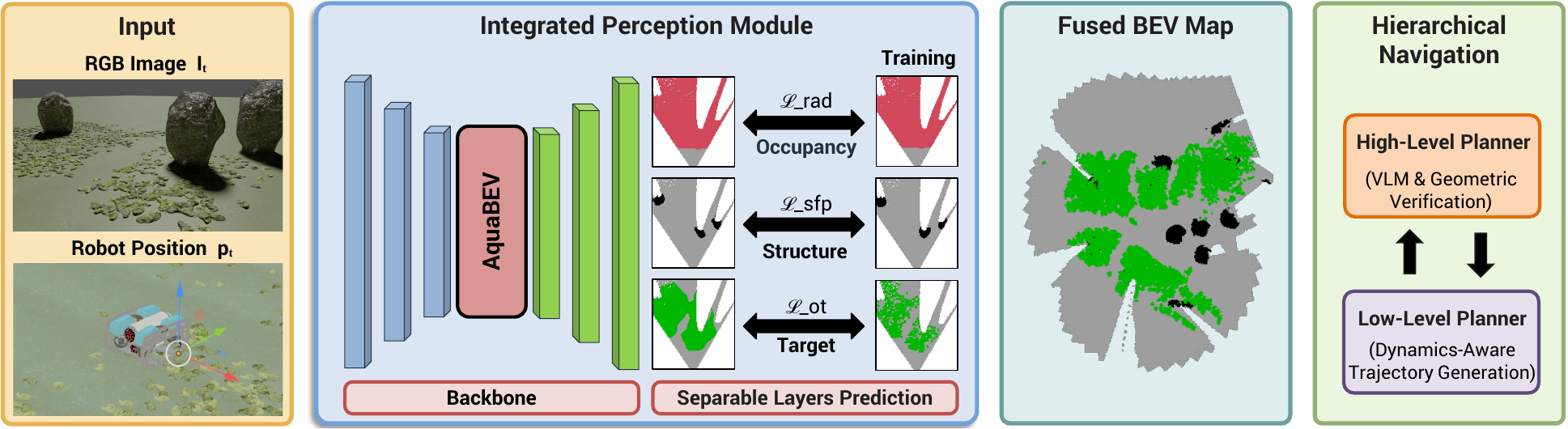}
    \caption{\textbf{AquaBEV-Nav architecture.}
    AquaBEV-Nav predicts local structure, target, and observability fields from a
    single RGB image. These fields are transformed into the world frame and
    accumulated in navigation system persistent spatial memory.}
    \label{fig:arch}
    \vspace{-3mm}
\end{figure*}

\subsection{Planner Ready BEV Prediction}

At time $t$, AquaBEV-Nav receives an RGB image $I_t$ and vehicle pose
$T_t \in SE(3)$. Instead of estimating monocular depth, reconstructing a point
cloud, and rasterizing the result into occupancy, AquaBEV-Nav directly predicts
the spatial representation consumed by the navigation system.

We retain the AquaBEV perception trunk~\cite{dong2026aquabev}. Image features
are mapped through learned polar queries into a feature field
$F(\theta,r)$, processed causally along the range dimension, and reconstructed
into Cartesian BEV occupancy. This preserves the direct RGB to BEV formulation of
AquaBEV while adapting its output to the requirements of persistent navigation.

A binary occupancy field alone is insufficient for navigation systems because different
parts of the navigation stack query different spatial information. We therefore
predict three independent fields, 
\begin{equation}
f_{\theta}(I_t)
=
(\hat{s}_t,\hat{g}_t,\hat{o}_t),
\end{equation}
where $\hat{s}_t$ represents standing structure, $\hat{g}_t$ represents mission
targets, and $\hat{o}_t$ represents observability. Independent sigmoid outputs
allow these fields to overlap rather than forcing each cell into a single
semantic class. This is important when, for example, a mission target lies on
an otherwise navigable seabed. Structure is routed to the obstacle map, target
predictions to the semantic map, and observability to the explored area map.

\subsection{Planner Aware Spatial Objectives}
\label{sec:losses}

Cellwise supervision treats spatial errors independently of how they affect the
planner. AquaBEV-Nav therefore introduces field specific objectives that target
the dominant geometric errors of each output, as shown in Fig.~\ref{fig:arch}. Radial transport and structure
distillation act on the structure field, while spatial false positive and
target transport losses act on the target field.

All variants retain an occupancy preservation term,
\begin{equation}
\mathcal{L}_{\mathrm{anchor}}
=
\left\|
z_{\mathrm{occ}}
-
z_{\mathrm{occ}}^{\mathrm{frozen}}
\right\|_2^2 ,
\end{equation}
where $z_{\mathrm{occ}}$ is the current occupancy logit and
$z_{\mathrm{occ}}^{\mathrm{frozen}}$ is produced by a frozen copy of the
original AquaBEV model. This constrains adaptation of the planner specific
fields without allowing the underlying occupancy representation to drift.

\textbf{Radial transport.}
For structure, the distance of an obstacle along its viewing direction is
directly relevant to collision constraints. We therefore penalize radial
displacement using the one dimensional Earth Mover distance. Let
$p_{\theta j}$ and $q_{\theta j}$ denote predicted and ground truth structure
mass at range bin $j$. Each valid ray is first normalized,
\begin{equation}
\tilde{p}_{\theta j}
=
\frac{p_{\theta j}}
{\sum_j p_{\theta j}+\epsilon},
\qquad
\tilde{q}_{\theta j}
=
\frac{q_{\theta j}}
{\sum_j q_{\theta j}+\epsilon}.
\end{equation}
The radial objective is then
\begin{equation}
\mathcal{L}_{\mathrm{rad}}
=
\frac{1}{|\Theta^{+}|}
\sum_{\theta\in\Theta^{+}}
\Delta r
\sum_{k=1}^{N_r}
\left|
\sum_{j\leq k}\tilde{p}_{\theta j}
-
\sum_{j\leq k}\tilde{q}_{\theta j}
\right|,
\end{equation}
where $\Theta^{+}$ contains rays with valid structure supervision.
Normalizing each ray makes the objective sensitive to spatial displacement
rather than differences in total predicted mass.

\textbf{Spatial target extent.}
Target predictions require a different spatial constraint. Excess target mass
can spread beyond the true object and distort the semantic region later used
for target localization. We therefore weight false positive target mass by its
distance $D_i$ from the nearest true target cell,
\begin{equation}
\mathcal{L}_{\mathrm{sfp}}
=
\frac{
\sum_i p_i^g(1-y_i^g)\min(D_i,D_{\max})
}{
\sum_i p_i^g(1-y_i^g)+\epsilon
}\delta ,
\end{equation}
where $p_i^g$ and $y_i^g$ denote predicted and ground truth target values and
$\delta$ is the BEV cell size. Nearby errors receive a smaller penalty, while
target mass extending farther from the true spatial support is penalized more
strongly.

\textbf{Target transport.}
Controlling target extent does not ensure that the predicted target is centered
at the correct location. We therefore additionally compare normalized target
distributions along both range and azimuth using
$\mathcal{L}_{\mathrm{tgt}}$. Radial displacement is already expressed in
metres. Angular displacement is converted to metres at a $5$ m reference range,
placing the two transport components in a common unit. Since both distributions
are normalized before transport, this objective measures target displacement
and is intentionally independent of total predicted target mass.

\textbf{Structure distillation.}
A second RayOccNet receives ground truth depth during training and acts as a
frozen teacher for the RGB student. Rather than distilling depth itself, we
transfer the teacher's structure representation in polar space. Let
$z^s_{\theta r}$ and $z^t_{\theta r}$ denote student and teacher structure
logits. We use
\begin{equation}
\mathcal{L}_{\mathrm{distill}}
=
\frac{1}{N}
\sum_{\theta,r}
\left(1+\frac{r}{5}\right)
\left(
z^s_{\theta r}
-
z^t_{\theta r}
\right)^2 .
\end{equation}
The weighting places greater emphasis on distant structure, where monocular
visual geometry is weakest. The teacher is used only during training and is
discarded at inference.

The complete objective is
\begin{equation}
\mathcal{L}
=
\mathcal{L}_{\mathrm{task}}
+
\lambda_a\mathcal{L}_{\mathrm{anchor}}
+
\lambda_r\mathcal{L}_{\mathrm{rad}}
+
\lambda_s\mathcal{L}_{\mathrm{sfp}}
+
\lambda_t\mathcal{L}_{\mathrm{tgt}}
+
\lambda_d\mathcal{L}_{\mathrm{distill}},
\end{equation}
with $\lambda_a=1.0$ and the remaining weights selected using validation data.

\begin{table*}[t]
\centering
\caption{\textbf{Planner Facing BEV Prediction Performance.} Comparison of occupancy, structure, and target prediction under a unified
training protocol. All learned models are initialized from their respective
real world checkpoints, anchored to frozen occupancy teachers, and trained for
25 epochs under the same schedule.(\textbf{BEST}/ \underline{SECOND BEST})}
\label{tab:perception}
\footnotesize
\setlength{\tabcolsep}{4pt}
\begin{tabular}{l|ccc|ccc|ccc}
\toprule
& \multicolumn{3}{c|}{Occupancy} & \multicolumn{3}{c|}{Structure $\rightarrow$ \texttt{grid}} & \multicolumn{3}{c}{Target $\rightarrow$ \texttt{seg.\ map}} \\
\cmidrule(lr){2-4}\cmidrule(lr){5-7}\cmidrule(lr){8-10}
Arm & mIoU $\uparrow$ & Prec. \,\% $\uparrow$ & Recall\,\% $\uparrow$ & mIoU $\uparrow$ & Prec.\,\% $\uparrow$ & Recall\,\% $\uparrow$ & mIoU $\uparrow$ & Prec.\,\% $\uparrow$ & Recall\,\% $\uparrow$ \\
\midrule
\emph{CORAL (DAv2)}~\cite{wu2026coral}      & ---  & ---  & ---  & 10.0 & 10.6 & 65.2 & 11.0 & 18.3 & 21.5 \\
\midrule
TPVFormer~\cite{huang2023tpvformer}               & 14.8 & 76.1 & 15.6 & 11.0 & 22.2 & 18.0 & 38.6 & 45.8 & 71.0 \\
SurroundOcc~\cite{wei2023surroundocc}             & 16.7 & 75.3 & 17.7 & 16.0 & 31.7 & 24.5 & 38.9 & 47.4 & 68.3 \\
MonoScene~\cite{cao2022monoscene}                 & 24.1 & 80.4 & 25.6 & 21.5 & 26.8 & 52.4 & 39.7 & 43.5 & 82.0 \\
GaussianFormer-fix~\cite{huang2024gaussianformer} & 23.5 & 80.0 & 24.9 & 27.6 & 31.0 & \textbf{71.5} & 41.8 & 47.0 & 78.8 \\
AquaBEV~\cite{dong2026aquabev}                    & \underline{28.3} & \underline{83.6} & \underline{29.9} & \underline{29.9} & \underline{41.5} & 51.7 & \underline{46.6} & \underline{51.5} & \textbf{83.1} \\
\midrule
\textbf{AquaBEV-Nav (ours)}                       & \textbf{30.5} & \textbf{84.1} & \textbf{32.4} & \textbf{37.48} & \textbf{55.3} & \underline{56.3} & \textbf{53.2} & \textbf{59.4} & \underline{79.6} \\

\bottomrule
\end{tabular}
\vspace{-4mm}
\end{table*}

\begin{table*}[t]
\centering
\caption{\textbf{Accumulated planner map performance.}
Comparison of structure and target maps after trajectory level fusion. (\textbf{BEST}/ \underline{SECOND BEST})}
\label{tab:mapspace}
\footnotesize
\setlength{\tabcolsep}{2.5pt}
\begin{tabular}{l|cccc|cccc}
\toprule
& \multicolumn{4}{c|}{\textbf{Structure}} & \multicolumn{4}{c}{\textbf{Target}} \\
\cmidrule(lr){2-5}\cmidrule(lr){6-9}
Arm  & max IoU $\uparrow$ & max P $\uparrow$ & mean IoU $\uparrow$ & mean P $\uparrow$ 
     & max IoU $\uparrow$ & max P $\uparrow$ & mean IoU $\uparrow$ & mean P $\uparrow$  \\
\midrule
\emph{CORAL (DAv2)}~\cite{wu2026coral}       & \phantom{0}8.3 & \phantom{0}8.3 & 21.5 & 25.7   & 25.0 & 28.7 & \phantom{0}8.5 & 20.7  \\
\midrule
TPVFormer~\cite{huang2023tpvformer}                & 21.9 & 25.1 & \phantom{0}7.9 & 47.7   & 47.5 & 48.8 & 47.8 & 67.7  \\
SurroundOcc~\cite{wei2023surroundocc}              & 27.4 & 32.0 & 11.8 & 71.7  & 47.6 & 49.1 & 46.2 & 71.8 \\
MonoScene~\cite{cao2022monoscene}                  & 22.5 & 23.7 & 31.3 & 56.7  & 46.0 & 46.6 & 53.2 & 63.0  \\
GaussianFormer-fix~\cite{huang2024gaussianformer}  & 27.1 & 28.0 & \textbf{44.7} & 55.4  & 52.2 & 53.8 & 53.8 & 67.5  \\
AquaBEV~\cite{dong2026aquabev}                     & \underline{33.8} & \underline{37.6} & 37.8 & 80.8   & \underline{55.4} & \underline{56.5} & \underline{58.3} & \underline{75.6}  \\
\midrule
\textbf{AquaBEV-Nav (ours)}                        & \textbf{45.2} & \textbf{51.9} & \underline{41.4} & \textbf{86.0}   & \textbf{67.9} & \textbf{71.6} & \textbf{59.8} & \textbf{85.0}  \\
\bottomrule
\end{tabular}
\vspace{-3mm}
\end{table*}

\subsection{Integration with CORAL}
\label{sec:integration}

At each navigation step, the three local BEV fields are transformed into the
world frame using $T_t$ and accumulated into CORAL's persistent maps. Structure
updates obstacles, target updates semantic memory, and observability updates
explored space. All downstream components, including target tracking, vision
language reasoning, geometric verification, and local trajectory generation,
remain unchanged from CORAL~\cite{wu2026coral}. This isolates the effect of the
perception representation on closed loop navigation.
\vspace{-1mm}
\section{Experiments}
\label{sec:exp}
\vspace{-2mm}
\subsection{Setup}

\textbf{Simulation.}
\textbf{Simulation.}
We use CORAL's Blender reef environments~\cite{wu2026coral}, comprising
16 runs across ten environments and five reef topologies. The dataset contains
2,945 RGB frames with semantic renders, vehicle poses, and BEV labels. We split
complete environments into train $\{s_1,s_3,s_4,s_6,s_7,s_9\}$ with 1,978
frames, validation $\{s_2,s_8\}$ with 494 frames, and test
$\{s_5,s_{10}\}$ with 473 frames. Runs from the same environment remain within
the same split. Generated BEV labels agree with an independent reference at
$98.1$, $96.2$, and $99.9$ IoU for observability, target, and structure.
For closed loop evaluation, all perception front ends use the same simulator
segmentation to control target semantics. CORAL constructs its spatial map from
Depth Anything V2~\cite{yang2024depthanythingv2}, whereas AquaBEV-Nav predicts
the BEV representation directly from RGB. All downstream mapping, reasoning,
planning, and control components remain unchanged.

\textbf{Real world deployment.}
Physical trials use a BlueROV2 Heavy equipped with a front-facing camera
and a Waterlinked Sonar3D-15 in a controlled swimming pool. The sonar
provides the obstacle map used by CORAL's local planner but is not an input
to the AquaBEV-Nav perception network. Cups serve as mission targets and
floating balloons as obstacles. AquaBEV-Nav is initialized from uScenes
weights~\cite{dong2026uscenes} and adapted using the simulation data.

\textbf{Metrics and protocol.}
Perception is evaluated at three levels, per frame IoU, precision, recall, and
predicted to ground truth mass; accumulated world maps under max and mean
fusion; and spatial error decomposition by extent, range, and hallucination.
Navigation uses CORAL's original metrics~\cite{wu2026coral}, coverage,
coverage time, collisions, VLM calls, and verifier
deviations. All six perception arms train for 25 epochs at three paired seeds.
We report the mean from epochs 20 through 24. A fixed checkpoint bootstrap over
2,000 frame resamples gives noise floors of $\pm0.84$ structure IoU and
$\pm0.50$ target IoU.


\begin{figure*}[!t]
    \centering
    \includegraphics[width=\textwidth, trim=0 2.5cm 0 0, clip]
    {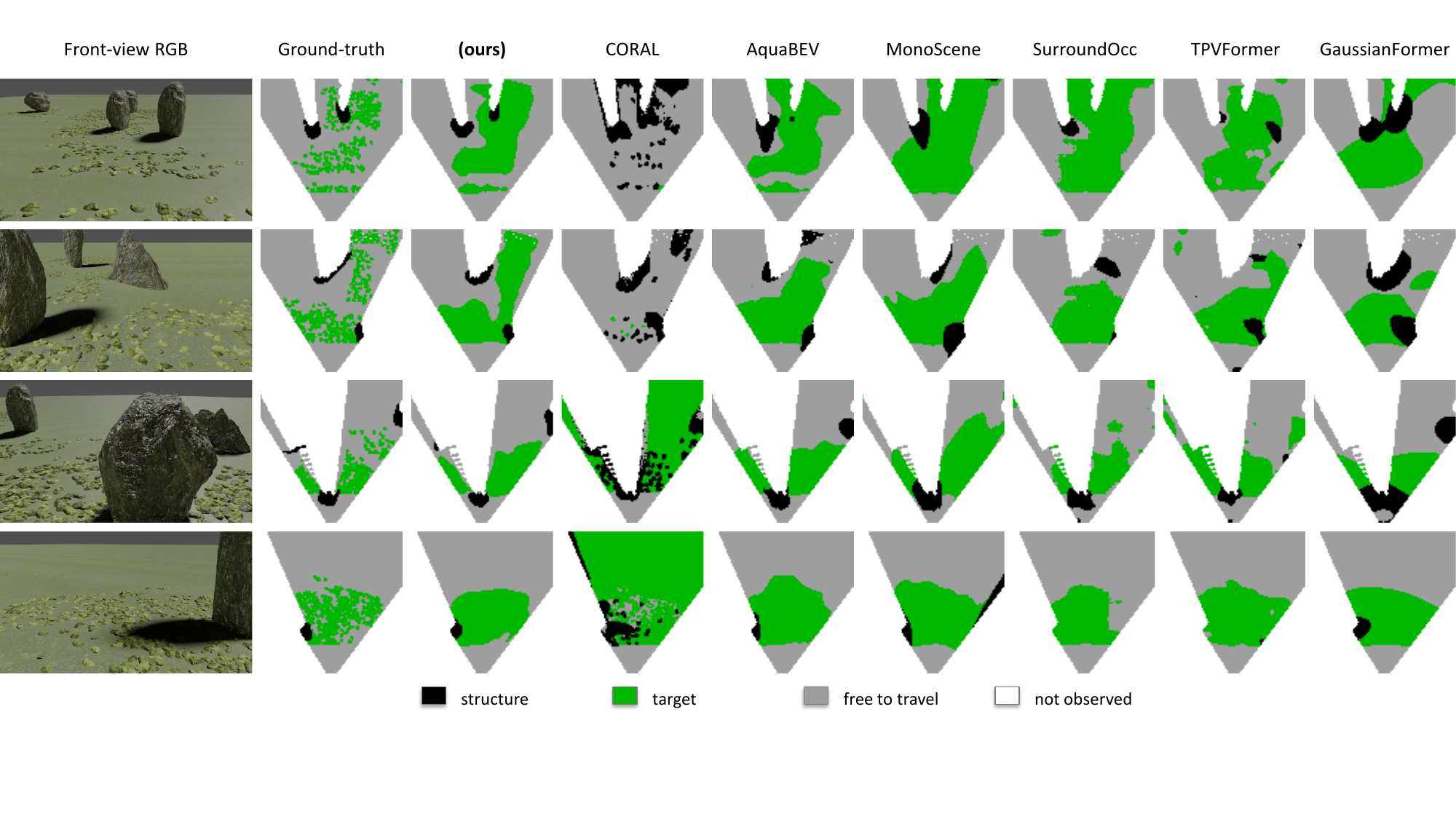}
    \vspace{-1mm}
    \caption{\textbf{Simulation perception.}
    Qualitative comparison of local BEV predictions across representative
    environments.}
    \label{fig:qualitative}
    \vspace{-4mm}
\end{figure*}

\vspace{-1mm}
\subsection{Simulation}
\label{sec:sim}

\subsubsection{Perception and Mapping}

Table~\ref{tab:perception} reports occupancy and planner field prediction under
the same protocol. Occupancy recall spans $15.6$ to $32.4\,\%$ across the
learned arms. AquaBEV-Nav reaches $30.5$ occupancy IoU, compared with $28.3$
for the AquaBEV baseline. The planner fields separate the models more strongly.
Structure occupies only $2.08\,\%$ of observed training cells, while target
occupies $14.7\,\%$, and target IoU ranges from $38.6$ to $53.2$ across the
learned arms.

Per frame accuracy does not fully determine the map seen by the planner.
Table~\ref{tab:mapspace} therefore accumulates predictions along each validation
trajectory. Max fusion improves target IoU for every learned arm by $6.3$ to
$14.7$ points, while structure improves only when placement remains consistent
across observations. AquaBEV increases from $29.9$ to $33.8$ structure IoU,
whereas GaussianFormer decreases from $27.6$ to $27.1$. Mean fusion changes
the ordering again, motivating the joint reporting of IoU, precision, and
predicted mass.

\begin{table}[t]
\centering
\caption{\textbf{Architecture ablation for planner field prediction.}Comparison of structural changes applied to the AquaBEV backbone.}
\vspace{-1mm}
\label{tab:arch}
\footnotesize
\setlength{\tabcolsep}{4pt}
\begin{tabular}{@{}l |cc |cc@{}}
\toprule
\multirow{2}{*}{\textbf{Method}}
& \multicolumn{2}{c|}{\textbf{Structure}} & \multicolumn{2}{c}{\textbf{Target}} \\
\cmidrule(lr){2-3}\cmidrule(lr){4-5}
 & \textbf{mIoU$\uparrow$} & \textbf{Prec. \%$\uparrow$ }& \textbf{mIoU$\uparrow$} & \textbf{Prec. \%$\uparrow$}\\
\midrule
AquaBEV
  & 29.47\,$\pm$\,0.31 & 39.0\,$\pm$\,1.8
  & 46.39\,$\pm$\,0.50 & 51.9\,$\pm$\,1.5 \\
\;+ ray PE
  & 29.28\,$\pm$\,0.76 & 39.1\,$\pm$\,2.0
  & 45.94\,$\pm$\,0.50 & 51.2\,$\pm$\,1.3 \\
\;+ ray consist.
  & 29.24\,$\pm$\,0.55 & 39.3\,$\pm$\,2.7
  & 45.91\,$\pm$\,0.25 & 50.5\,$\pm$\,0.8 \\
\;+ ray PE \& consist.
  & 29.30\,$\pm$\,0.07 & 38.5\,$\pm$\,0.3
  & 45.81\,$\pm$\,0.91 & 50.8\,$\pm$\,2.3 \\
\;+ range CE
  & \textbf{30.35\,$\pm$\,0.41} & \textbf{39.8\,$\pm$\,1.4}
  & \textbf{47.10\,$\pm$\,0.11} & \textbf{52.1\,$\pm$\,0.9} \\
\bottomrule
\end{tabular}
\vspace{-4mm}
\end{table}

\begin{table}[t]
\centering
\caption{\textbf{Planner aware objective ablation.}Effect of the proposed field specific training objectives on structure and
target prediction.}
\vspace{-1mm}
\label{tab:loss-ablation}
\footnotesize
\setlength{\tabcolsep}{4pt}
\begin{tabular}{@{}l |cc|cc@{}}
\toprule
\multirow{2}{*}{\textbf{Method}}
& \multicolumn{2}{c|}{\textbf{Structure}} & \multicolumn{2}{c}{\textbf{Target}} \\
\cmidrule(lr){2-3}\cmidrule(lr){4-5}
 & \textbf{mIoU$\uparrow$} & \textbf{Prec. \%$\uparrow$ }& \textbf{mIoU$\uparrow$} & \textbf{Prec. \%$\uparrow$} \\
\midrule
AquaBEV
  & 29.47\,$\pm$\,0.31 & 39.8\,$\pm$\,1.4
  & 46.39\,$\pm$\,0.50 & 51.9\,$\pm$\,1.5 \\
\;+ $\mathcal{L}_{\mathrm{rad}}$
  & 36.00\,$\pm$\,0.71 & 49.5\,$\pm$\,1.7
  & 46.32\,$\pm$\,0.65 & 50.4\,$\pm$\,1.5 \\
\;+ $\mathcal{L}_{\mathrm{sfp}}$
  & 35.96\,$\pm$\,0.76 & 49.9\,$\pm$\,2.9
  & 48.54\,$\pm$\,0.38 & 53.6\,$\pm$\,1.5 \\
\;+ $\mathcal{L}_{\mathrm{ot}}$
  & 36.26\,$\pm$\,0.56 & 51.7\,$\pm$\,4.1
  & \textbf{52.74\,$\pm$\,0.20} & \textbf{59.6\,$\pm$\,1.0} \\
\;+ distillation
  & \textbf{37.39\,$\pm$\,0.62} & \textbf{54.6\,$\pm$\,0.4}
  & 51.74\,$\pm$\,0.29 & 59.1\,$\pm$\,0.2 \\
\bottomrule
\end{tabular}
\vspace{-6mm}
\end{table}

\begin{figure*}[!t]
    \centering
    \includegraphics[width=\textwidth]
    {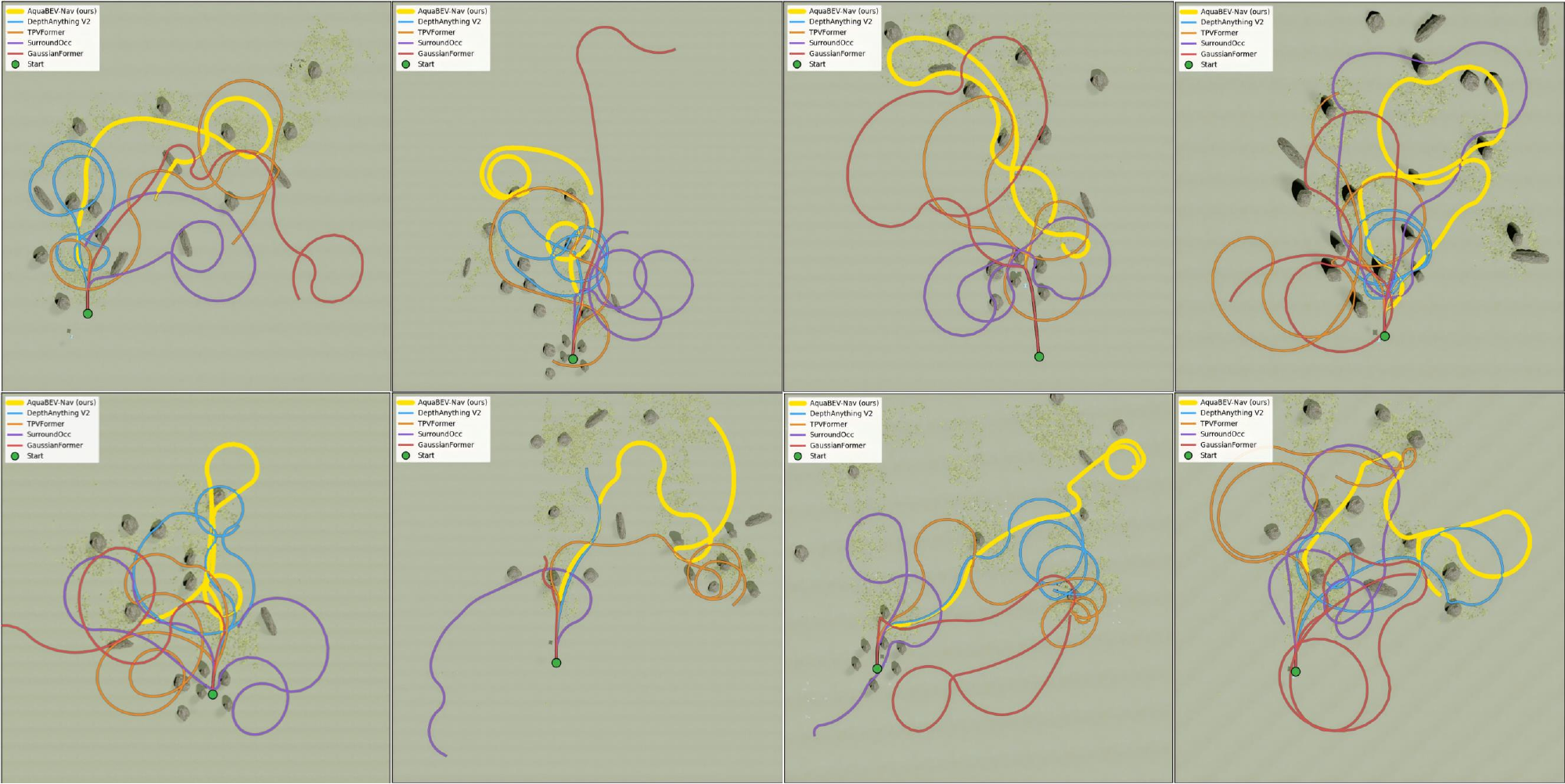}
    \vspace{-1mm}
    \caption{\textbf{Closed loop navigation by reef topology.}
    Executed trajectories produced by different perception front ends
under the same CORAL navigation stack.}
    \label{fig:sim-nav}
    \vspace{-4mm}
\end{figure*}

The error decomposition further shows that structure false positives are
dominated by range error, accounting for $53$ to $70\,\%$, whereas target
false positives are dominated by lateral extent at $48$ to $78\,\%$. In the
far field, $55$ of $63$ ground truth structures have a prediction on the
correct bearing, indicating that range placement rather than visibility is a
major source of error. GaussianFormer achieves a $0.10$ m same bearing range
error at matched mass despite weaker IoU, further showing that overlap alone
does not describe map quality.

\subsubsection{Ablations}

Across 17 architectural modifications, Table~\ref{tab:arch} reports the
representative variants evaluated under the final protocol. The auxiliary range
classification objective is the only displayed modification that improves both
planner fields beyond the measured noise floor, increasing structure IoU from
$29.47$ to $30.35$ and target IoU from $46.39$ to $47.10$, corresponding to
gains of $0.88$ and $0.71$. Its associated lifting gate converges to
$\alpha=0.00000$ at every seed, indicating that the gain comes from range
supervision rather than the additional lifting path. Other changes to feature
transport, positional encoding, and range reasoning do not improve both fields
consistently. This suggests that the principal limitation is not insufficient
architectural capacity, but the spatial supervision used to shape the learned
representation. These results redirect the final design toward stronger spatial
supervision rather than additional topology changes.

We first verify that the occupancy anchor is necessary during semantic
adaptation: removing it reduces occupancy IoU from $37.9$ to $20.4$, showing
that the adapted trunk can otherwise overwrite the geometry inherited from
AquaBEV. Table~\ref{tab:loss-ablation} then evaluates the planner aware
objectives cumulatively. Radial transport primarily improves structure,
reaching $36.00$ IoU while leaving target performance nearly unchanged.
Adding the target specific spatial false positive loss changes structure by
only $-0.04$ while improving target IoU by $2.22$. Target transport further
raises target IoU from $48.54$ to $52.74$, while structure distillation produces
the strongest structure result at $37.39$ IoU. The resulting improvements, shown in Fig. \ref{fig:qualitative}, are
also reflected qualitatively in the sharper structure placement and more compact
target predictions.

\begin{table}[t]
\centering
\caption{\textbf{Closed-loop simulation navigation comparison across ten reef environments.} }
\label{tab:nav-sim}
\footnotesize
\setlength{\tabcolsep}{2.5pt}
\begin{tabular}{@{}l@{\hspace{3pt}}cccccc@{}}
\toprule
Front end & cov.\% $\uparrow$ & time\,s $\downarrow$ & coll.\ $\downarrow$ & VLM $\downarrow$ & dev.\ $\downarrow$ \\
\midrule
DREAM~\cite{wu2025dream} & 80.00 & 1513.2  & \phantom{0}9 & 1261 & 15 \\
CORAL (DA2)  & 59.58 & 620  & \phantom{0}0 & \phantom{0}519 & \phantom{0}3 \\
\midrule
\;+ TPVFormer~\cite{huang2023tpvformer} & 68.33 & 734   & 0 & 191 & 19 \\
\;+ GaussianFormer~\cite{huang2024gaussianformer}\!\! & 43.21 & 612  & 0 & 394 & 19 \\
\;+ SurroundOcc~\cite{wei2023surroundocc}  & 49.17 & 669 & 0 & 188 & 29 \\
\;+ \textbf{AquaBEV-Nav (ours)}                & 88.95 & 734 & 0 & 353 & 5 \\
\midrule
\emph{CORAL}~\cite{wu2026coral}  & 94.28 & \phantom{0}642.0  & \phantom{0}0 & \phantom{0}547 & \phantom{0}3 \\
\bottomrule
\end{tabular}
\vspace{-4mm}
\end{table}

\subsubsection{Navigation}
Table~\ref{tab:nav-sim} evaluates the full CORAL exploration loop with perception as the only changing component. AquaBEV-Nav achieves the highest coverage among the learned front ends at $88.95\%$, compared with $59.58\%$ for the DAv2-based CORAL front end and $68.33\%$ for the strongest learned baseline, TPVFormer. Although TPVFormer and SurroundOcc invoke the VLM less frequently, their substantially lower coverage indicates that fewer queries alone do not imply more effective exploration. AquaBEV-Nav also produces only five verifier-detected deviations, substantially fewer than the 19--29 deviations observed for the other learned BEV front ends, suggesting that its accumulated spatial representation provides more stable guidance to the downstream planner. The published CORAL system reports $94.28\%$ coverage under its original
experimental configuration~\cite{wu2026coral}, providing a reference point
for the current controlled evaluation.

Fig.~\ref{fig:sim-nav} complements these aggregate results by showing the executed trajectories across representative reef topologies. The different front ends produce noticeably different exploration patterns even though the downstream CORAL stack is unchanged. In the representative cases, the yellow AquaBEV-Nav trajectories more closely follow the spatial distribution of the oyster targets, allowing the vehicle to progress continuously along target regions and cover them with fewer unnecessary detours. In contrast, several other front ends exhibit more pronounced
circuitous or repeated motions before reaching the same regions. The figure further illustrates how perception quality affects which regions are revisited, which branches are explored, and how the robot progresses through more constrained layouts. Together, Table~\ref{tab:nav-sim} and Fig.~\ref{fig:sim-nav} show that improvements in the learned BEV representation translate beyond per-frame perception accuracy into more complete and stable closed-loop exploration.

\subsection{Real World Deployment}
\label{sec:real}

Pool evaluation is qualitative because no instrumented BEV ground truth or
target cluster annotation is available. We therefore compare predicted geometry
against synchronized sonar and inspect accumulated maps and executed
trajectories against the known physical layout. 

\begin{figure}[!t]
    \centering
    \includegraphics[width=\textwidth, trim=2cm 1.5cm 3cm 0, clip]
    {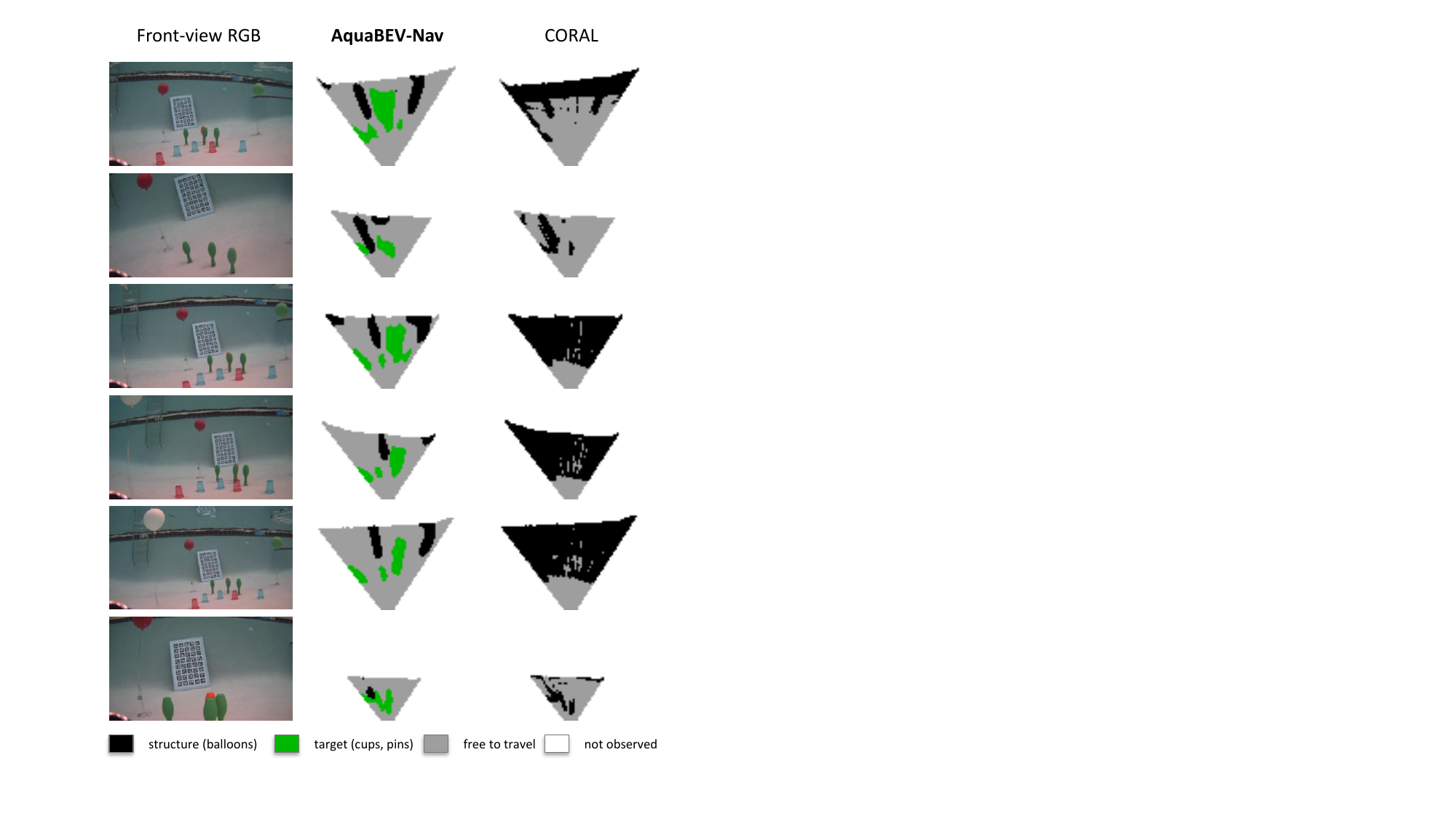}
    \caption{\textbf{Pool perception.} Representative front-view RGB observations
and corresponding local BEV predictions from AquaBEV-Nav and the
depth-based CORAL front end.}
    \label{fig:real-perc}
    \vspace{-3mm}
\end{figure}

Observability also exposes a deployment issue hidden by simulation. A static
sensor wedge reaches $65.7$ IoU but marks approximately $2,200$ cells per frame
as observed despite no corresponding measurement, which incorrectly expands the
explored region behind occluding structure. Estimates derived from the auxiliary
range head perform substantially worse, reaching only $18.6$ IoU by expectation
and $35.3$ by argmax, because a single range estimate cannot recover the full
spatial pattern of visibility. AquaBEV-Nav therefore predicts observability
directly as a separate field, allowing explored space to vary with scene
geometry rather than with a fixed sensor footprint. The resulting local
structure predictions and accumulated map are shown in
Fig.~\ref{fig:real-perc}.

\section{Conclusion}

We presented \textbf{AquaBEV-Nav}, a framework that connects learned monocular
BEV occupancy with hierarchical underwater navigation. Rather than constructing
the planner map through an intermediate monocular depth estimate, AquaBEV-Nav
predicts the required spatial representation directly from RGB and integrates it
into CORAL's persistent memory while leaving the downstream reasoning, planning,
and control stack unchanged. This formulation allows the effect of the learned
BEV representation to be evaluated independently from changes to the navigation
system itself.

\subsection{Discussion}

Our results show that strong occupancy prediction alone does not guarantee a
representation suitable for navigation. CORAL requires distinct information
about physical structure, mission targets, and explored space, motivating the
decomposition of learned occupancy into separate structure, target, and
observability fields. This distinction is important because the same occupied
region can have different meanings for perception and planning. A returned
surface may correspond to navigable terrain, an obstacle, or a mission target,
and these cases should not be represented by a single binary field.

The experiments further show that spatial supervision matters more than
additional architectural complexity. Across the architectural variants tested,
only auxiliary range supervision consistently improves both planner fields,
whereas the planner aware objectives produce substantially larger changes in the
specific spatial errors they are designed to address. Radial transport improves
structure placement along range, while target specific objectives reduce target
displacement and excessive spatial extent. Structure distillation provides
additional geometric supervision where monocular prediction remains difficult.
Together, these results suggest that learned occupancy for navigation should be
optimized according to how spatial errors affect the planner rather than only
according to aggregate overlap accuracy.

\subsection{Future Work}

The current study evaluates AquaBEV-Nav in simulated reef environments and
controlled pool trials. Future work should extend evaluation to larger and more
diverse natural underwater environments containing greater variation in
visibility, terrain, vehicle motion, and target appearance. Longer deployments
will also make it possible to study how local prediction errors accumulate in
persistent spatial memory over extended trajectories.

A second direction is to expose prediction uncertainty directly to the planner.
The current experiments intentionally keep CORAL fixed in order to isolate the
effect of the perception representation. Future systems could instead allow the
planner to reason over confidence in structure, target, and observability
predictions and adjust exploration or collision constraints accordingly. More
generally, jointly optimizing learned spatial perception and downstream planning
offers a promising direction for improving closed loop underwater autonomy while
preserving a modular interface between perception and navigation.

\balance
\bibliographystyle{IEEEtran}
\bibliography{ref}

\end{document}